\pdfoutput=1

\documentclass[11pt]{article}

\usepackage[]{acl}

\usepackage{times}
\usepackage{latexsym}
\usepackage[T1]{fontenc}
\usepackage[utf8]{inputenc}

\usepackage{microtype}

\usepackage{inconsolata}

\usepackage{graphicx}
\usepackage{array}
\usepackage{amsmath}
\usepackage{multirow}
\usepackage{booktabs}
\usepackage{subcaption}
\usepackage{xcolor}
\usepackage{arydshln} 
\usepackage{bbding}
\usepackage{algpseudocode}  
\usepackage[ruled,vlined,linesnumbered]{algorithm2e}

\title{Beyond Guilt: Legal Judgment Prediction with Trichotomous Reasoning}

\author{
    Kepu Zhang\textsuperscript{1},
    Haoyue Yang\textsuperscript{2},
    Xu Tang\textsuperscript{2},
    Weijie Yu\textsuperscript{2}\thanks{~Corresponding author.},
    Jun Xu\textsuperscript{1} \\
    \textsuperscript{1}Gaoling School of Artificial Intelligence, Renmin University of China \\
    \textsuperscript{2} University of International Business and Economics \\
    \texttt{kepuzhang@ruc.edu.cn } 
   \\
}

\begin{document}
\maketitle

\begin{abstract}
In legal practice, judges apply the \textit{trichotomous dogmatics of criminal law}\footnote{Translated from the German legal term ``der trichotomisch aufgebauten Dogmatik des Strafrechts''.}, sequentially assessing \textit{the elements of the offense}, \textit{unlawfulness}, and \textit{culpability}\footnote{Translated from the German legal terms ``Tatbestand'', ``Rechtswidrigkeit'', and ``Schuld''.} to determine whether an individual's conduct constitutes a crime. Although current legal large language models (LLMs) 
show promising accuracy in judgment prediction,
they lack trichotomous reasoning capabilities due to the absence of an appropriate benchmark dataset, preventing them from predicting ``innocent''
\footnote{In this paper, the terms "non-guilty'' and ``innocent'' are used interchangeably.} 
outcomes.
As a result, every input is automatically assigned a charge, limiting their practical utility in legal contexts.
To bridge this gap, we introduce LJPIV, the first benchmark dataset for \underline{L}egal \underline{J}udgment \underline{P}rediction with \underline{I}nnocent \underline{V}erdicts. Adhering to the trichotomous dogmatics, we extend three widely-used legal datasets through LLM-based augmentation and manual verification. Our experiments with state-of-the-art legal LLMs and novel strategies that integrate trichotomous reasoning into zero-shot prompting and fine-tuning reveal: (1) current legal LLMs have significant room for improvement, with even the best models achieving an F1 score of less than 0.3 on LJPIV; and (2) our strategies notably enhance both in-domain and cross-domain judgment prediction accuracy, especially for cases resulting in an innocent verdict.
\end{abstract}

\section{Introduction}
The trichotomous dogmatics~\cite{elias2015three,dubber2005theories} is a key theory in criminal law used to determine whether an individual's conduct constitutes a crime. It is widely applied in civil law jurisdictions such as Germany, Japan, and China. In legal practice, judges apply this framework in three sequential steps: (1) the elements of the offense, where the individual's conduct is assessed to see if it objectively fulfills the criteria for a criminal offense; (2) unlawfulness, where it is determined whether the individual has any grounds for justification—such as self-defense—that would exempt them from criminal liability despite fulfilling the offense’s elements; and (3) culpability, which requires that the individual acted with free will to be held responsible. If the individual acted while mentally ill or lacking judgment, they are deemed to lack criminal responsibility and thus cannot be considered guilty. As illustrated in Figure 1, through trichotomous reasoning, the conduct must satisfy the elements of the offense, unlawfulness, and culpability in order to constitute a crime.
\begin{figure}[t]
    \centering
    \includegraphics[width=\linewidth]{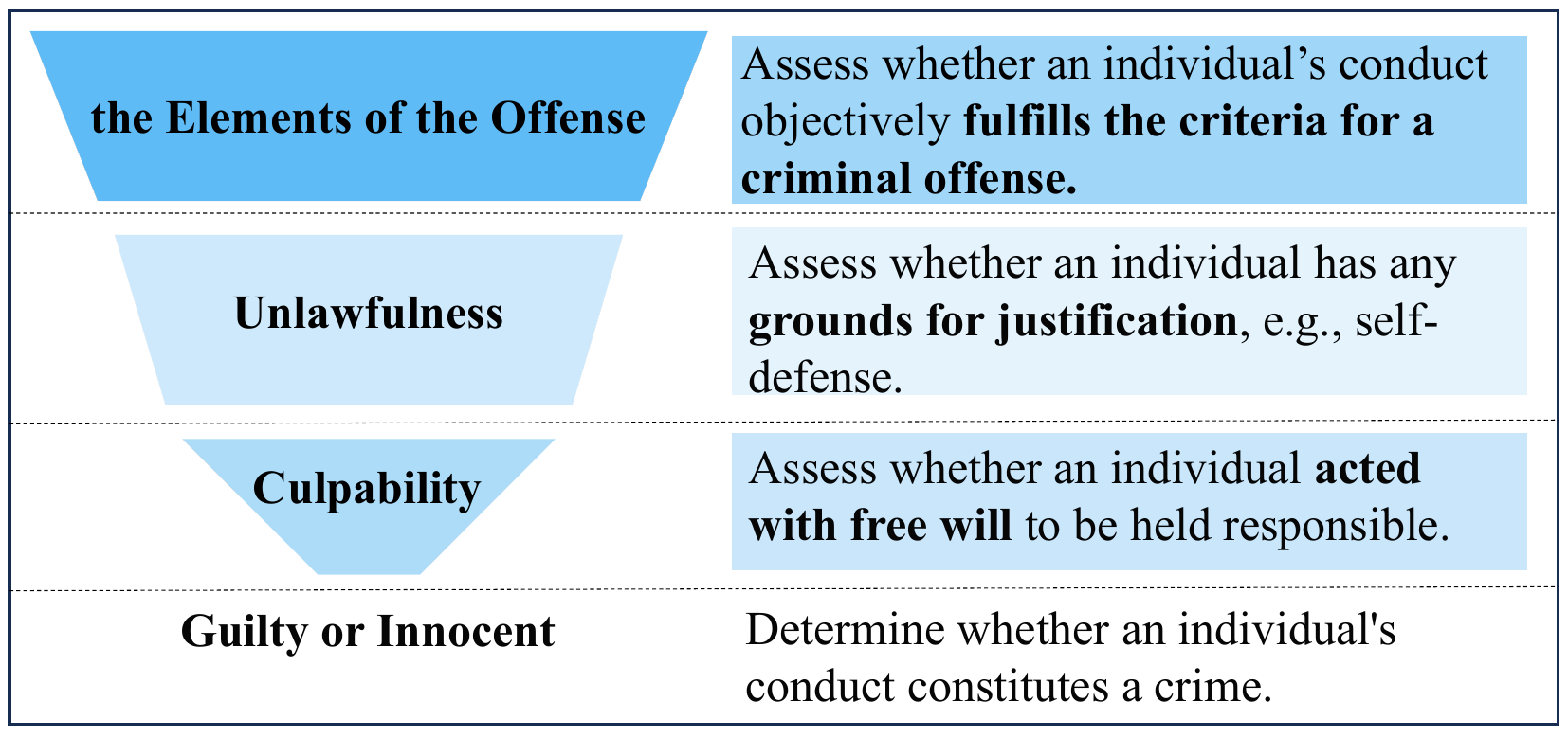}
    \caption{
    An illustration of trichotomous dogmatics of criminal law. The reasoning process proceeds sequentially from top to bottom. Conduct that satisfies the elements of the offense, unlawfulness, and culpability is deemed to constitute a crime; otherwise, the individual is considered innocent.
    }
    \label{fig:three}
    \vspace{-3mm}
\end{figure}

\begin{figure*}
    \centering
\includegraphics[width=0.98\linewidth]{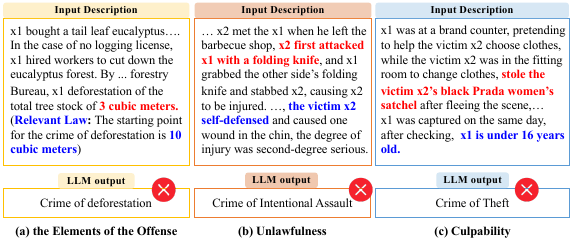}
    \caption{
    DISC-Law~\cite{yue2023disc} incorrectly predicts charges for non-guilty fact descriptions across the elements of the offense, unlawfulness, and culpability. The \textcolor{red}{red} parts represent actions that may lead to a guilty verdict, while the \textcolor{blue}{blue} parts indicate acts or situations that result in contradictions or exoneration.
    }
    \label{fig:fig1}
    \vspace{-3mm}
\end{figure*}
In the field of legal NLP~\cite{sdu_fuzi_mingcha,cui2023chatlaw,shu2024lawllm}, researchers fine-tune large language models (LLMs) on large-scale legal judgment prediction (LJP) corpora to enable them to predict applicable criminal charges based on an input fact description. While these models~\cite{yue2023disc,sdu_fuzi_mingcha} have achieved impressive accuracy in charge prediction, they only map an individual’s conduct to the elements of a criminal offense—the first step in trichotomous dogmatics—and thus lack the full trichotomous reasoning capability. As a result, a major limitation of existing legal LLMs is that they automatically assign a charge to the input without the ability to predict an ``innocent'' outcome, even when there are clear grounds for justification. For instance, as shown in Figure 2, DISC-Law~\cite{yue2023disc}, the state-of-the-art Chinese legal LLM, overlooks self-defense in example 2(b) and an age requirement for criminal responsibility in 2(c), leading to incorrect crime predictions. This limitation significantly reduces their practical utility in legal contexts. The primary cause of this issue is the lack of a benchmark dataset designed to support trichotomous reasoning in legal judgment prediction.

To address this gap, we introduce LJPIV, the first benchmark dataset for \underline{L}egal \underline{J}udgment \underline{P}rediction with \underline{I}nnocent \underline{V}erdicts. Unlike previous benchmarks that only provide fact descriptions and corresponding charge labels, LJPIV focuses on trichotomous reasoning, particularly for cases with innocent outcomes, by offering fine-grained, sentence-level labels that indicate compliance with the trichotomous dogmatics of criminal law. To construct LJPIV, we extend three popular benchmarks—CAIL~\cite{xiao2018cail2018,shui2023comprehensive}, ELAM~\cite{yu2022explainable}, and LeCaRD~\cite{ma2021lecard,deng2024element}—through a three-stage augmentation process.
\textit{First}, recognizing that some sentences in the fact descriptions (such as suspect profiles) may be less informative for judicial decisions, we fine-tune an LLM to extract sentences relevant to trichotomous reasoning, reducing the risk of over-correction~\cite{li2023effectiveness,fang2023chatgpt} in the subsequent stages.
\textit{Second}, we apply the retrieval-augmented generation (RAG)~\cite{lewis2020retrieval,gao2023retrieval} technique to retrieve relevant criminal laws for the input and prompt the LLM to introduce grounds for justification—legal reasons that exempt the individual from criminal liability despite their conduct meeting the elements of the offense. This allows us to randomly select a portion of the data and create counterfactual samples\footnote{Please note that current LJP datasets are derived from publicly available cases, all of which involve guilty verdicts. Therefore, we refer to augmented samples corresponding to not-guilty situations as counterfactual samples.} labeled as ``innocent''.
\textit{Third}, to ensure the logical consistency and coherence of these counterfactual samples, we prompt the LLM to perform a self-check, followed by multiple rounds of manual verification by annotators.

We further introduce zero-shot prompt-based and fine-tuning methods to equip open-domain LLMs~\cite{qwen} with trichotomous reasoning capabilities, particularly for predicting innocent outcomes in legal judgment prediction. Extensive experiments on state-of-the-art legal LLMs reveal that (1) current legal LLMs have significant room for improvement, with even the best models achieving an F1 score below 0.3 on LJPIV; (2) fine-tuning on LJPIV substantially improves both in-domain and cross-domain judgment prediction accuracy for open-domain LLMs, especially in cases resulting in an innocent verdict; and (3) our trichotomous reasoning strategies including the prompt-based and fine-tuning approaches further enhance the legal judgment performance.

Our contributions can be summarized as follows:
\begin{itemize}
\item We identify a significant limitation of current legal LLMs: their inability to predict ``Innocent'' outcomes for given fact descriptions, which restricts their practical utility in legal contexts. Inspired by the trichotomous dogmatics of criminal law, our work pioneers the integration of trichotomous reasoning capabilities into LLMs for legal judgment prediction, particularly for predicting innocent outcomes.
\item We construct the first benchmark dataset for legal judgment prediction with innocent verdicts by extending three popular benchmarks through a three-stage LLM-based augmentation process, followed by manual verification.
\item Extensive experiments demonstrate that fine-tuning on LJPIV significantly enhances both in-domain and cross-domain judgment prediction accuracy for open-domain LLMs, particularly for cases with innocent verdicts, while also validating the effectiveness of our trichotomous reasoning strategies in further improving judgment performance.
\end{itemize}

\section{Related Work}
\subsection{Legal Judgment Prediction}
\begin{table}[t]
\centering
\resizebox{\columnwidth}{!}{
\begin{tabular}{lccc}
    \toprule
    
    \textbf{\shortstack{Dataset}}  &\textbf{\shortstack{LLM \\ Annotation}}
    &\textbf{\shortstack{Non-guilt \\ Label}}&\textbf{\shortstack{Trichotomous \\ Reasoning}}
    \\
    \midrule
    \textbf{CAIL-2018~\cite{xiao2018cail2018}} &\XSolidBrush&\XSolidBrush&\XSolidBrush\\
    \textbf{CAIL-Long~\cite{xiao2021lawformer}} &\XSolidBrush&\XSolidBrush&\XSolidBrush\\
    \textbf{ELAM~\cite{yu2022explainable}} &\XSolidBrush&\XSolidBrush&\XSolidBrush\\
    \textbf{LeCaRD~\cite{ma2021lecard}} &\XSolidBrush&\XSolidBrush&\XSolidBrush\\
    \textbf{DPAM~\cite{wang2018modeling}} &\XSolidBrush&\XSolidBrush&\XSolidBrush\\    
    \textbf{SLJA~\cite{deng2023syllogistic}} &\CheckmarkBold&\XSolidBrush&\XSolidBrush\\
    \midrule
    \textbf{LJPIV (Ours)}&\CheckmarkBold&\CheckmarkBold&\CheckmarkBold\\
    \bottomrule
\end{tabular}
}
\caption{Comparison between existing LJP datasets with our LJPIV. LJPIV is the only one that includes not-guilty labels and supports trichotomous reasoning.
}
\label{tab:ljp_dataset}
\vspace{-3mm}
\end{table}

Legal judgment prediction is a classic legal task, aiming to predict a charge based on the case facts. Prior to the LLMs era, legal judgment prediction models evolved through stages including those based on statistical rules~\cite{segal1984predicting,gardner1987artificial,nagel1963applying}, small models (such as BERT)~\cite{chalkidis2019neural,zhong2018legal}, reinforcement learning~\cite{lyu2022improving,zhao2022charge}, and the integration of legal domain knowledge~\cite{wu2023precedent,zhang2023case}.
We review the datasets for legal judgment prediction, as summarized in Table~\ref{tab:ljp_dataset}. The CAIL-2018~\cite{xiao2018cail2018} initiated the task of Chinese legal judgment prediction.
ELAM~\cite{yu2022explainable} and LeCaRD~\cite{ma2021lecard}, have been proposed for judicial case matching tasks, where each case includes a corresponding case description and charge. These datasets have been used in legal judgment prediction tasks~\cite{sun2024logic,qin2024explicitly}.
These datasets primarily rely on manual annotation, requiring the hiring of professional legal workers and consuming significant time and resources.
SLJA~\cite{deng2023syllogistic} is a legal judgment prediction dataset derived using syllogistic reasoning. 
Differing from these datasets, this paper introduces datasets built around a smaller-scale LLM using trichotomous reasoning that includes not-guilty cases.

\subsection{LLM in Legal NLP}
With the rapid development of LLMs, their performance in various open-domain tasks~\cite{peng2023instruction,achiam2023gpt} has been remarkable. The exploration of LLM applications in legal NLP tasks is burgeoning~\cite{fei2023lawbench,choi2021chatgpt}. By incorporating legal reasoning steps into the prompts~\cite{blair2023can,yu2022legal} of LLMs, these models are guided to complete specific legal tasks. Additionally, techniques such as Retriever-Augmented Generation (RAG)~\cite{zhang2023reformulating,zhou2023boosting,pipitone2024legalbench} are utilized to retrieve relevant case law and content from legal knowledge bases to assist LLMs in completing legal tasks. Furthermore, by fine-tuning open-domain LLMs with extensive legal documents and legal task datasets, a surge in specialized legal LLMs~\cite{yue2023disc,cui2023chatlaw,shu2024lawllm}, has been observed.
The objective of this paper is to utilize LLMs to annotate a dataset that includes not-guilty cases and to explore how LLMs can implement trichotomous reasoning.

\section{Dataset Construction}
We focus on trichotomous reasoning for the legal judgment prediction task, which can be formulated as follows: given an input fact description $x$, the goal is to predict its judgment outcome $y\in\mathcal{Y}$. Unlike existing studies where $\mathcal{Y}$ represents a set of criminal charges, we consider a more practical scenario by extending $\mathcal{Y}$ to include a label for ``innocent''. Specifically, we utilize LLMs to augment three popular LJP datasets and create a new benchmark, LJPIV, through a three-stage augmentation.

\subsection{Sentence Extraction for Reasoning}
When utilizing LLMs to extend legal datasets, over-modification~\cite{li2023effectiveness,fang2023chatgpt} must be carefully considered, as it is unacceptable for legal documents due to their serious nature; excessive modifications can alter the legal meaning of the text and negatively impact trichotomous reasoning. Additionally, some sentences, such as suspect profiles, may be less informative for the judicial decisions. Therefore, in the first stage, we extract sentences that correspond specifically to trichotomous reasoning.

To achieve this goal, we construct an instruction fine-tuning dataset, $D_E$, using LeCaRD-Elem~\cite{deng2024element}, where the queries contain legal element annotations. Based on $D_E$, we develop an instruction template in the format $\{Instruction, (Case, Crime), Crime-related \ Sentence\}$ and use LoRA~\cite{hu2021lora} to efficiently fine-tune LLM~\cite{qwen}. This process allows us to inject the necessary knowledge for extracting reasoning-related sentences into the LLM, resulting in the sentence extractor $E_C$.

To extend existing LJP datasets, we segment them into sentences and apply $E_C$ to extract those relevant to trichotomous reasoning. Given the relatively small size of $D_E$ (approximately 100 entries), we also implement a decoding constraint~\cite{geng2023grammar,lu2024diver} on $E_C$ that adjusts the LLM’s probability distribution during decoding to ensure each sentence is extracted only once.

\begin{figure*}
    \centering
\includegraphics[width=0.98\linewidth]{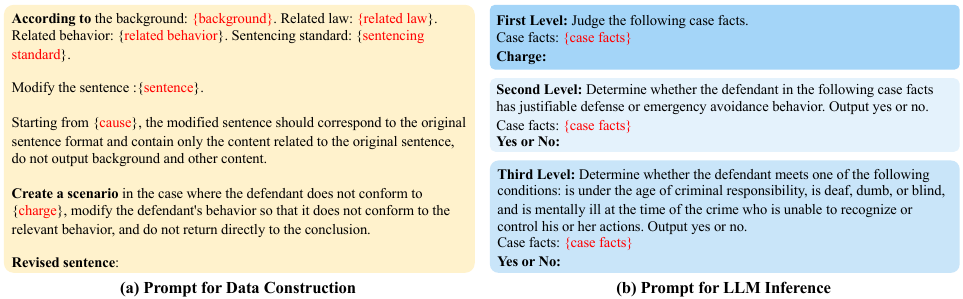}
    \caption{
    The prompt for trichotomous reasoning used in this study. 
    }
    \label{fig:prompt}
    \vspace{-3mm}
\end{figure*}

\subsection{Injection of Grounds for Justification}\label{sec:injection}
After extracting sentences relevant to trichotomous reasoning from the input fact descriptions, we randomly select 50\% samples from each dataset and leverage the LLM's strong semantic understanding and instruction-following capabilities~\cite{ouyang2022training,achiam2023gpt} to generate counterfactual samples, guiding the LLM to inject grounds for justification into the extracted sentences. We use carefully designed prompts as shown in Figure~\ref{fig:prompt}(a), within a RAG framework~\cite{lewis2020retrieval,gao2023retrieval} to introduce acts or situations that result in contradictions or exoneration at the levels of the elements of the offense, unlawfulness, and culpability, based on the retrieved criminal laws relevant to the case.

At the level of \textbf{the Elements of the Offense}, judges assess whether an individual’s conduct objectively fulfills the criteria for a criminal offense.
Considering that the current LJP datasets consist of already adjudicated cases, where each sample has assigned labels such as applicable charges and relevant legal articles, we derive exonerating actions and circumstances based on the legal definitions of the charges to construct counterfactual samples. Specifically, we use the charge as a query to retrieve the corresponding legal articles, criminal behaviors, and judgment criteria. Using these retrieval results, we guide the LLM to generate scenarios or actions that contradict the establishment of the charge and incorporate them into the original fact description to create not-guilty samples. For example, as shown in Figure~\ref{fig:fig1}(a), the legal definition for the crime of deforestation requires that the total volume of trees felled reaches 10 cubic meters. We instruct the LLM to modify the total volume of trees felled in the original case to below 10 cubic meters, such as 3 cubic meters, making the charge invalid.

At the level of \textbf{Unlawfulness}, judges assess whether an individual has any grounds for justification, meaning legal reasons that exempt a person from criminal liability despite fulfilling the elements of the offense. In Chinese criminal law, two primary situations lead to exoneration at the level of unlawfulness: self-defense and necessity. Therefore, we instruct the LLM to modify the original fact description by introducing a scenario involving either self-defense or necessity, so that the case satisfies the exoneration conditions for the charge. For example, as shown in Figure~\ref{fig:fig1}(b), although x2 caused second-degree harm to x1, his actions fall under the category of self-defense, meaning that x2 is not guilty of intentional injury. Please note that not all charges are applicable to self-defense or necessity. For instance, in the deforestation case shown in Figure~\ref{fig:fig1}(a), so we skip this step for such fact descriptions.

At the level of \textbf{Culpability}, judges assess whether an individual acted with free will and can be held responsible. In this study, we primarily consider three situations: (1) the defendant has not yet reached the age of criminal responsibility, (2) the defendant is deaf, mute, or blind, or (3) the defendant is a person with a mental illness who was unable to recognize or control their actions at the time of the crime. Since these situations are related to the description of the defendant, we randomly select one of these scenarios and instruct the LLM to incorporate it into the defendant’s profile. For example, as shown in Figure~\ref{fig:fig1}(c).

\subsection{Data Quality Verification}
Through manual inspection, we find that LLM-based augmentation can lead to over-modification. For example, the LLM may incorrectly treat a fracture as a minor injury, which contradicts common sense, and should instead modify it to a truly minor injury, such as a scratch. To address this logical inconsistency, we implement both an LLM self-check and manual verification to ensure the quality of our LJPIV dataset.

For the LLM self-check, we instruct LLM to check whether augmented cases, i.e. those counterfactual samples with non-guilty labels, have the aforementioned issues with the prompt \textit{``Determine whether the following case facts have logical problems, common sense errors, contradictions, unreasonable or incoherent content''.}

After the LLM's self-check, we employ five legal annotators to manually verify the correctness of the augmented samples from both legal and logical perspectives.
Specifically, we use a multi-round random inspection process. In each round, each annotator randomly selects 20\% of the augmented samples and compares the fact descriptions against the legal provisions to ensure the samples are consistent with legal innocence. After each round, five annotators collaboratively revise the over-modified samples based on legal standards. The revised ones are excluded from the next round of inspection. This process is repeated until no issues are found in the randomly selected cases. Ultimately, we conducted five rounds of inspection and corrections.

\section{Trichotomous Reasoning with LLM}
In this section, we introduce both prompt-based and fine-tuning methods to enable LLMs to perform trichotomous reasoning for LJP.

\subsection{Prompt-Based Method}
Given an input fact description $x$, with our carefully designed trichotomous prompts $p_1$, $p_2$, and $p_3$, as illustrated in Figure~\ref{fig:prompt} (b), the LLM generates three predictions $y_1$, $y_2$, and $y_3$. Each prediction corresponds to the reasoning outcome for the levels of the elements of the offense, unlawfulness, and culpability, respectively:
\begin{equation}
    y_k=f_{\mathrm{LLM}}(x,p_k;\theta),
\end{equation}
where $\theta$ represents the parameters of the LLM; $k\in\{1,2,3\}$ denotes the index for the reasoning level. Specifically, $y_1$ represents a charge or non-guilt prediction, while $y_2$ and $y_3$ are ``Yes/No'' responses from the LLM, indicating whether there are grounds for justification and whether the individual has criminal responsibility, respectively. 
Therefore, the overall judgment prediction $y_{\text{final}}$ can be derived as follows:
\begin{equation}
    y_{\text{final}}=
    \left\{\begin{array}{ll}
    y_1& \text { if } y_2=\text{``No''}, y_3=\text{``No''},  \\
    \text{non-guilt} & \text { otherwise}
    \end{array}\right.
\end{equation}

\subsection{Fine-Tuning-Based Method}
Fine-tuning the LLM using our LJPIV  in the following two steps is an alternative approach to equip the LLM with trichotomous reasoning capabilities.

\textbf{Fine-tuning Dataset $D_{NG}$ Construction}. 
For each level of trichotomous reasoning, the input for fine-tuning consists of the fact description and the corresponding prompt, as shown in Figure~\ref{fig:prompt}(b). In terms of the output, at the level of the elements of the offense, the output is either a criminal charge or an innocent label. At the levels of unlawfulness and culpability, the output is a ``Yes/No'' response, indicating whether there are grounds for justification and whether the individual has criminal responsibility, respectively.

\textbf{Fine-tuing and Inference}.
We perform fine-tuning using LoRA~\cite{hu2021lora} on $D_{NG}$ : 
\begin{equation}
\label{eq:loss_1}
    \mathcal{L}_{\mathrm{FT}} = - \frac{1}{|\mathcal{D}_{NG}|} \sum_{\mathcal{D}_{NG}} \mathrm{log}(P_{\theta+\theta_{L}}(y_t|x,p,y_{<t})),
\end{equation}
where $\theta$ and $\theta_{L}$ represent the parameters of the LLM and LoRA, respectively; $y_t$ denotes the $t$-th token, and $y_{<t}$ represents the tokens preceding $y_t$.
In the inference phase, we utilize the fine-tuned LLM to make the judgment in three sequential steps:
\begin{equation}
\label{eq:ft}
    y_k=f_{\mathrm{LLM}}(x,p_k;\theta+\theta_L),
\end{equation}
where $k \in \{1,2,3\}$ denotes the index for the reasoning level in trichotomous dogmatics.

\section{Experiments}
\subsection{Experimental Settings}
\subsubsection{Dataset and Metric}
\textbf{Dataset.} 
We conduct extensive experiments on the proposed LJPIV datasets, which are constructed by extending CAIL-2018~\cite{xiao2018cail2018}, ELAM~\cite{yu2022explainable}, and LeCaRD~\cite{ma2021lecard}, referred to as LJPIV-CAIL, LJPIV-ELAM, and LJPIV-LeCaRD, respectively. Each sample in LJPIV consists of a fact description and the corresponding judgment (non-guilty or a charge). The ratio of guilty to innocent samples is set to 1:1. Among the innocent samples, the reasons for innocence, corresponding to the three levels of trichotomous dogmatics, are balanced with a ratio of 3:1:1. Specific statistical information about these datasets is provided in Table~\ref{tab:data_main}.

Since both LLM-based and manual annotating are costly,
we follow~\cite{shui2023comprehensive} and select a subset of CAIL-2018, called CAIL-train, as the training set for LJPIV. The remainder of CAIL-2018, along with ELAM and LeCaRD, serve as in-domain and cross-domain test sets for LJPIV. In addition, since ELAM and LeCaRD are based on retrieved datasets where only the queries have corresponding crime labels, we limit the test set to include only queries to ensure the quality of the judgment labels. Annotating crime labels for the candidate cases is reserved for future work.
 
\begin{table}[t]
\centering
\resizebox{\columnwidth}{!}{
\begin{tabular}{lccccc}
    \toprule
    \multirow{1}{*}{\textbf{Dataset}} &\multirow{1}{*}{\#\textbf{Case}} &\multirow{1}{*}{\#\textbf{Charge}} &\multirow{1}{*}{\textbf{Avg\_Case\_Len}} 
    &\multirow{1}{*}{\textbf{Avg\_Sent\_Num}}\\
    \midrule
    \textbf{LJPIV-CAIL-Train} &1120&112&439.43&6.23 \\
    \textbf{LJPIV-CAIL-Test} &560&112&436.63&6.16\\
    \textbf{LJPIV-ELAM}&500&63&832.82&10.50 \\
    \textbf{LJPIV-LeCaRD}&80&30&448.34&7.25 \\
    \bottomrule
\end{tabular}}
\caption{Statistics of our LJPIV. \textbf{\#Case}, \textbf{\#Charge} denote the number of the cases and charges. \textbf{Avg\_Case\_Len} and \textbf{Avg\_Num\_Sent} represent the average length of the case and the average number of sentences.}
\label{tab:data_main}
\vspace{-3mm}
\end{table}

\textbf{Metric.}
Following~\cite{feng2022legal,zhong2018legal}, we evaluate the LJP results using widely-used metrics including  Accuracy (Acc), Precision (P), Recall (R), and F$_1$-Score (F$_1$).

\begin{table*}[h!]
    \vspace{-3px}
    \resizebox{1\textwidth}{!}{
         \begin{tabular}{ll|cccc|cccc|cccc}
          \hline
          \multicolumn{2}{c|}{Dataset} & \multicolumn{4}{c|}{LJPIV-CAIL (in-domain)} & \multicolumn{4}{c|}{LJPIV-ELAM (cross-domain)} &\multicolumn{4}{c}{LJPIV-LeCaRD (cross-domain)}\\
          \hline
          Category & Model & Acc & P & R & F$_1$ & Acc & P & R & F$_1$& Acc & P & R & F$_1$\\
          \hline
          \multicolumn{1}{c}{\multirow {3}{*}{\shortstack{Legal LLM}}} 
          &Disc-LawLLM&30.54&30.52&35.12&29.50&24.00&15.39&18.00&14.53&20.00&20.15&21.34&16.74\\
          &LexiLaw &18.21&20.85&18.51&18.02&15.20&13.71&13.73&11.69&13.75&10.45&16.67&10.43\\
          &fuzi.mingcha &19.46&25.16&20.83&20.72&13.40&13.07&13.54&11.16&10.00&9.23&9.96&7.59
          \\
          \hline
          \multicolumn{1}{c}{\multirow {7}{*}{\shortstack{Qwen2\\(Qwen2-7B-Instruct)}}} 
          &Zero-shot&29.82&29.25&30.27&27.14&22.40&18.61&17.49&15.88&22.50&21.23&27.78&21.58\\
          &Zero-shot-CoT&35.18&31.51&31.76&29.12&27.20&19.09&16.50&15.44&26.25&23.22&26.76&21.08\\
          &Zero-shot-Tri (Ours)&50.71&35.90&34.75&32.68&38.60&22.18&\underline{19.80}&\underline{18.46}&42.50&\textbf{32.30}&\textbf{35.69}&\textbf{31.68} \\
          \cdashline{2-14}
          &Few-shot-BM25&36.79&34.65&35.48&32.36&25.00&18.63&19.18&15.83&26.25&\underline{25.51}&\underline{28.51}&\underline{23.22}\\
          &Few-shot-SBERT&36.07&35.86&36.17&32.94&24.00&19.31&18.98&16.61&23.75&24.15&26.99&22.02\\
          & Fine-Tuing-Direct&\underline{82.68}&\underline{66.11}&\underline{60.73}&\underline{60.62}&\underline{58.20}&\underline{23.23}&19.68&18.40&\underline{52.50}&8.30&10.26&8.15\\
          & Fine-Tuing-Tri (Ours)&\textbf{86.96}&\textbf{69.41}&\textbf{68.83}&\textbf{67.42}&\textbf{68.00}&\textbf{27.02}&\textbf{25.13}&\textbf{23.53}&\textbf{56.25}&23.90&19.73&20.22\\
          \hline
          \multicolumn{1}{c}{\multirow {7}{*}{\shortstack{Baichuan2\\ (Baichuan2-7B-Chat)}}} 
            &Zero-shot&24.46&25.54&24.11&21.93&20.60&15.12&14.53&12.85&16.25&12.62&16.45&11.57\\
          &Zero-shot-CoT&27.68&29.28&22.33&23.48&23.00&17.54&15.69&14.55&17.50&12.00&11.76&9,27\\
          &Zero-shot-Tri (Ours) &46.25&32.72&31.34&29.04&23.20&18.18&16.79&15.00&22.50&\underline{17.81}&16.41&13.53\\
          \cdashline{2-14}
          &Few-shot-BM25&27.68&29.28&22.33&23.48&23.20&16.29&15.35&13.46&25.00&13.41&21.45&13.77\\
          &Few-shot-SBERT&35.18&29.69&27.21&26.00&25.00&17.11&14.97&13.20&23.75&14.45&\underline{21.89}&\underline{14.92}\\
          & Fine-Tuing-Direct&\underline{82.32}&\underline{67.41}&\underline{60.77}&\underline{61.37}&\underline{62.00}&\underline{25.10}&\underline{20.86}&\underline{19.48}&\underline{52.50}&7.35&8.84&7.89\\
          & Fine-Tuing-Tri (Ours)&\textbf{87.32}&\textbf{70.76}&\textbf{68.08}&\textbf{66.76}&\textbf{65.00}&\textbf{28.22}&\textbf{23.58}&\textbf{22.32}&\textbf{63.75}&\textbf{27.95}&\textbf{23.19}&\textbf{23.63}\\
          \hline
        \end{tabular}
    }
        \caption{Performance comparisons between Tri (Trichotomous) and the baselines on LJPIV-CAIL, LJPIV-ELAM and LJPIV-LeCaRD datasets. The best performance is indicated in bold, and the second best is underlined. 
        }
    \vspace{-3mm}
    \label{tab:main results}
\end{table*}

\subsubsection{Baseline}
(1) \textbf{Legal LLMs}: We selected three popular legal LLMs for comparison: \textbf{DISC-LawLLM}~\cite{yue2023disc}, \textbf{LexiLaw}\footnote{https://github.com/CSHaitao/LexiLaw}, and \textbf{fuzi.mingcha}~\cite{sdu_fuzi_mingcha}. These models are fine-tuned on a large amount of legal NLP task data and legal-related documents to inject legal knowledge, including legal judgment prediction datasets. 
For instance, DISC-LawLLM specifically used a legal judgment prediction dataset of 27K for fine-tuning. 

(2) \textbf{Open-domain LLMs}: The legal LLMs mentioned above are based on earlier versions of open-domain LLMs, while the current open-domain LLMs have undergone several updates and now possess enhanced text analysis and generation capabilities, which in turn improve their ability to make predictions.
We select Qwen~\cite{qwen} and Baichuan~\cite{yang2023baichuan} as open-domain LLMs for our experiments, which have strong performers across various fields. 
We tested their judgment prediction abilities using the following methods:
\textbf{Zero-shot}: We directly asked the LLMs to predict convictions. 
\textbf{Zero-shot-CoT}: While asking the LLMs to predict, we specifically instructed them to pay attention to innocence.
\textbf{Few-shot-BM25}: We use 
BM25~\cite{robertson1995okapi} to retrieve similar cases and their corresponding charges to provide context for the LLMs when making predictions.
\textbf{Few-shot-SBERT}: We use 
Sentence-BERT (SBERT)~\cite{reimers2019sentence} to retrieve similar cases and their corresponding charges. 
The retrieval corpus is conducted using LJPIV-CAIL-Train dataset.
Additionally, we directly fine-tuned the LLMs using the LJPIV-CAIL-Train, which we refer to as \textbf{Fine-Tuning-Direct}.
For the implementation of trichotomous reasoning, we denote the methods as \textbf{Zero-shot-Tri} and \textbf{Fine-Tuning-Tri}, respectively. 

The link and license for the aforementioned datasets and LLMs can be found in Appendix~\ref{sec:appendix dataset}.

\subsubsection{Implementation Details}
For the specific selection of LLMs, we chose Qwen2-7B-Instruct and Baichuan2-7B-Chat. Our implementation utilizes Huggingface Transformers~\cite{wolf2020transformers} in the PyTorch framework. Considering the long length of legal documents and the consumption of computational resources, we chose to retrieve one example for few-shot retrieval. For Fine-Tuning-Tri, we used LoRA~\cite{hu2021lora} to efficiently fine-tune the LLMs. We used the Adam optimizer~\cite{kingma2014adam}, set the initial learning rate to 5e-5, batch size to 16, used a cosine learning rate schedule, and fine-tuned for three epochs. All experiments are conducted on Nvidia A6000 GPUs.

\subsection{Main Results}

We conducted legal judgment prediction experiments on three datasets, and the results are shown in Table~\ref{tab:main results}. From the table, we can draw the following conclusions:

\noindent$\bullet$ \textbf{Effectiveness of the Trichotomous Reasoning}. We found that across the three datasets and two open-domain base LLMs, the reasoning-based methods, Zero-shot-Tri and Fine-Tuning-Tri, consistently achieved encouraging results. This demonstrates the effectiveness of trichotomous reasoning in predicting convictions, as it enables the LLMs to consider both guilt and innocence, leading to more accurate predictions of innocence.

\noindent$\bullet$ \textbf{The Legal LLMs Perform Poorly}. 
It can be found that although the legal LLMs have been fine-tuned with a substantial amount of legal knowledge, the performance of the three legal LLMs in predicting verdicts is relatively poor, even inferior to the zero-shot performance of open-domain LLMs. This is because the legal LLMs were fine-tuned on datasets with guilty legal judgments only and have not been exposed to cases of innocence. Therefore, when given a case, they do not recognize the possibility of rendering a not-guilty verdict.

\noindent$\bullet$ \textbf{Cross Domain Results}. We can observe that when transferring the LLM fine-tuned on the LJPIV-CAIL dataset to the other two datasets, the improvement is not as significant as the improvement on the CAIL dataset itself. 
This is due to the different sources of cases in different datasets, which may result in variations in length and style (for instance, as shown in Table~\ref{tab:data_main}, the case length in ELAM is twice that of CAIL). The differences between datasets lead to variations in the effectiveness of judgment prediction.
However, it is noticeable that compared to direct fine-tuning, the fine-tuning method based on the trichotomous reasoning still achieves better results. This is because the second and third levels of the trichotomous reasoning, which consider the innocence of the case, have universality: different cases might be acquitted for similar reasons due to the nature of the unlawfulness and the culpability involved.

\subsection{Ablation Study on Trichotomous Levels}

\begin{table}[t]
    
    \vspace{-3px}
    \resizebox{\columnwidth}{!}{
         \begin{tabular}{ll|cccc}
          \hline
          Category & Model & Acc & P & R & F$_1$ \\
          \hline
          \multicolumn{1}{c}{\multirow {3}{*}{\shortstack{Qwen2}}} 
          &Tri &\textbf{86.96}&\textbf{69.41}&\textbf{68.83}&\textbf{67.42}\\
          &\ \ w/o Level 3&77.86&59.83&61.01&58.70 \\
           &\quad w/o Level 2 &69.29&57.53&59.21&56.64\\

          \hline
          \multicolumn{1}{c}{\multirow {3}{*}{\shortstack{Baichuan2}}} 
          &Tri&\textbf{87.32}&\textbf{70.76}&\textbf{68.08}&\textbf{66.76} \\
           &\ \ w/o Level 3 &78.21&62.83&61.22&59.04 \\
           &\quad w/o Level 2&68.75&59.91&60.42&57.44 \\
          \hline
        \end{tabular}
    }
    \caption{
    Ablation studies for trichotomous reasoning on LJPIV-CAIL.  Tri refers to Fine-Tuing-Tri. 
        }
    \vspace{-3mm}
    \label{tab:aba three}
\end{table}

To explore the effectiveness of each level in the trichotomous reasoning, we investigated the performance of two LLMs on the LJPIV-CAIL test set by progressively removing the third and second levels from the Fine-Tuning-Tri. The results are shown in Table~\ref{tab:aba three}, and we provide a detailed analysis below:

\noindent$\bullet$ \textbf{w/o Level 3}. This indicates the removal of the level of culpability from the entire reasoning process, which means not considering the three situations mentioned in Sec.~\ref{sec:injection}
in rendering a not-guilty verdict. The performance drop observed in both LLMs highlights the importance of the third level. Allowing LLMs to assess the defendant's age and capacity for responsibility can refine the process of reaching not-guilty verdicts.

\noindent$\bullet$ \textbf{w/o Level 2}. This indicates the removal of the level of Unlawfulness on top of removing level 3, meaning not considering not-guilty verdicts due to the defendant's actions being a result of self-defense or necessity. The decrease in prediction results observed in both LLMs underlines the importance of the second level. Allowing LLMs to judge whether the defendant's actions were justified by self-defense or necessity can also improve the accuracy of not-guilty verdicts.

These findings suggest that each level in the trichotomous reasoning has a significant role in improving the accuracy of legal judgment predictions. By incorporating considerations of the three levels in the trichotomous reasoning, LLMs are better equipped to handle the complexity of legal cases and provide more nuanced verdicts. 

\subsection{Guilty vs. Non-Guilty Predictions}
\begin{figure}[t]
    \centering
    \includegraphics[width=\linewidth]{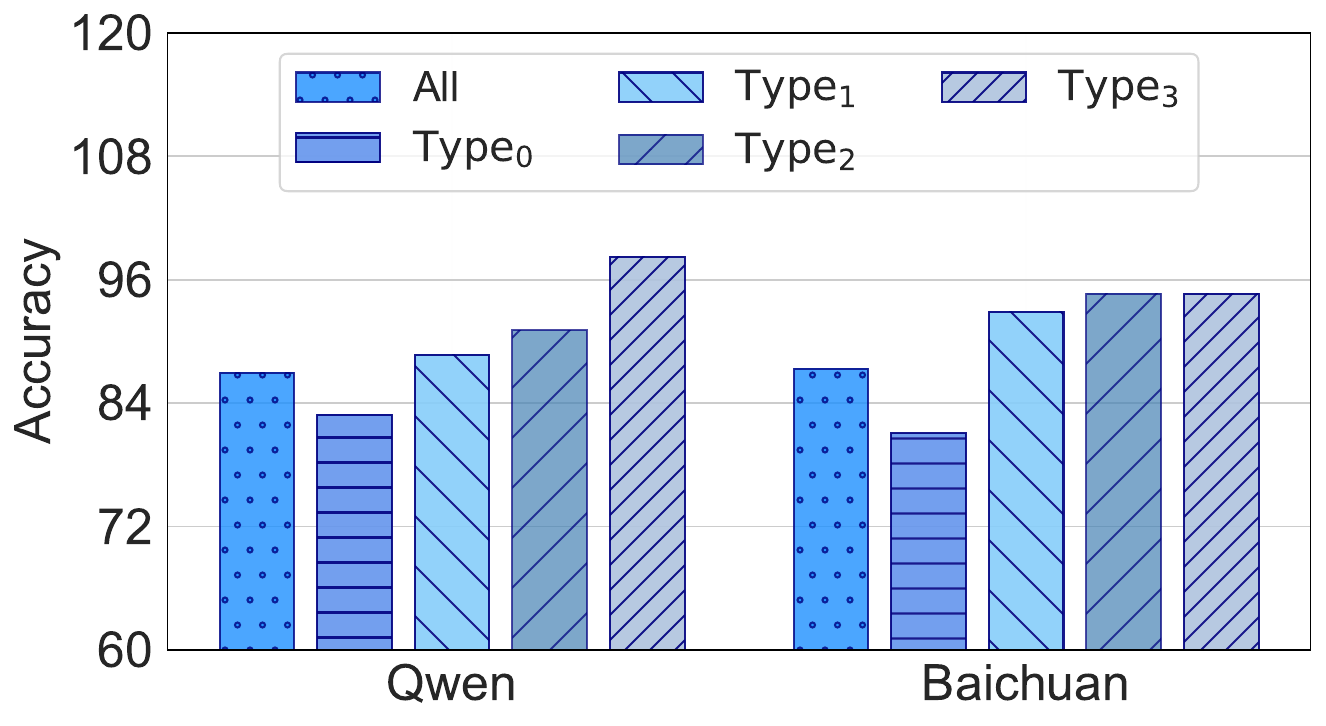}
    \caption{
    Prediction accuracy for different case types on the LJPIV-CAIL test set. ``All'' indicates the overall accuracy, while ``Type$_0$'' represents the accuracy for guilty cases. ``Type$_1$'', ``Type$_2$'', and ``Type$_3$'' represent the accuracies for non-guilty cases due to lack of elements, unlawfulness, and culpability, respectively. 
    }
    \label{fig:accuracy}
\end{figure}
The dataset contains both guilty and not guilty cases, with three types of not guilty verdicts: due to elements of offense, unlawfulness, and culpability. To evaluate how accurately the LLM predicts guilty and not guilty outcomes, we analyzed the final verdicts of two LLMs on the LJPIV-CAIL test set, categorizing them into four types: guilty (Type$_0$), not guilty due to elements of offense (Type$_1$), not guilty due to unlawfulness (Type$_2$), and not guilty due to culpability (Type$_3$). We then compared their predictive accuracies with the overall accuracy (All).

As shown in Figure~\ref{fig:accuracy}, both LLMs exhibited similar trends. The prediction accuracy for not guilty cases (Type$_1$, Type$_2$, Type$_3$) was higher than the overall accuracy (All), while the accuracy for guilty cases (Type$_0$) was lower. This suggests that the trichotomous reasoning is particularly effective for predicting not-guilty verdicts. When comparing the three types of not-guilty verdicts, we observed that the accuracy for cases not guilty due to elements was lower than those for unlawfulness and culpability. This difference arises because these characteristics vary in complexity. Elements require LLMs to differentiate between multiple reasons for not guilty verdicts across various charges, whereas unlawfulness and culpability are more generalizable.

Additionally, we analyzed the performance of the best-performing legal LLM, DISC-LawLLM, on the CAIL test set. We found that its accuracy for guilty verdicts was 61.07\%, while the accuracy for all three types of not-guilty verdicts was 0\%. This is likely due to overfitting during fine-tuning, causing the model to favor guilty verdicts for given cases, as shown in Figure~\ref{fig:fig1}.

\section{Conclusion}
In this paper, we introduce LJPIV, the first benchmark dataset for legal judgment prediction with innocent verdicts. We extend three widely-used legal datasets through LLM-based augmentation and manual verification. We further introduce zero-shot prompt-based and fine-tuning methods to equip open-domain LLMs with trichotomous reasoning capabilities, particularly for predicting innocent outcomes in legal judgment prediction. Extensive experiments reveal that (1) current legal LLMs have significant room for improvement, with even the best models achieving an F1 score below 0.3 on LJPIV; (2) fine-tuning on LJPIV substantially improves both in-domain and cross-domain judgment prediction accuracy for open-domain LLMs, especially in cases resulting in an innocent verdict; and (3) our trichotomous reasoning strategies further enhance the legal judgment performance.

\section{Limitations}
Legal systems across the world vary significantly, and different systems often adhere to distinct doctrines when making legal judgments. For instance, common law systems rely heavily on case precedents, whereas civil law systems are based on codified statutes. As a result, this paper focuses solely on Chinese datasets within the civil law system. Future work will aim to adapt our approach to other legal systems, including common law jurisdictions, and explore datasets in other languages to increase the generalizability and applicability of our approach across different legal contexts. Additionally, due to hardware resource constraints, we have only tested our methods on two open-domain LLMs, each with 7B parameters. In future work, we plan to experiment with larger-scale models and domain-specific legal LLMs to further enhance prediction accuracy and applicability across more complex legal tasks.

\section{Ethical Considerations}
The proposed LJPIV dataset is derived by extending existing LJP datasets, which are collected from officially publicly available sources and have been anonymized to protect the identities of the individuals involved. During the data augmentation process, to mitigate potential issues of over-correction by the LLM, we implemented a multi-step quality assurance process. This involves prompting the LLM to conduct a self-check to identify inconsistencies, followed by manual verification by human annotators to ensure the accuracy and integrity of the generated data.

\bibliography{custom}


\appendix

\section{More Details of Datasets and Models}\label{sec:appendix dataset}
In this section, we provide the link and license for the dataset we used, as shown in Table~\ref{tab:dataset urls and license}.

\begin{table*}[h]
\centering
\resizebox{0.9\textwidth}{!}{
\begin{tabular}{cl|ll}
    \toprule
    Type &Dataset &URL & Licence   \\
    \hline
    \multicolumn{1}{c}{\multirow {3}{*}{\shortstack{Dataset}}}
    &CAIL-2018 &\url{https://github.com/china-ai-law-challenge/CAIL2018}&MIT License\\ 
    &LeCaRD &\url{https://github.com/myx666/LeCaRD} &MIT License\\
    &ELAM &\url{https://github.com/ruc-wjyu/IOT-Match}&MIT License \\
    \hline
    \multicolumn{1}{c}{\multirow {5}{*}{\shortstack{LLM}}}
    &LexiLaw&\url{https://github.com/CSHaitao/LexiLaw}&MIT license\\
    &DISC-LawLLM &\url{https://github.com/FudanDISC/DISC-LawLLM}&Apache-2.0 license \\
    &fuzi.mingcha&\url{https://github.com/irlab-sdu/fuzi.mingcha} &Apache-2.0 license\\
    &Qwen2-7B-Instruct &\url{https://huggingface.co/Qwen/Qwen2-7B-Instruct}&Apache-2.0 license\\ 
    &Baichuan2-7B-Chat&\url{https://huggingface.co/baichuan-inc/Baichuan2-7B-Chat}&Apache-2.0 license\\ 
    \bottomrule
\end{tabular}
}
\caption{
The URLs and licenses for the datasets and LLMs used by LJPIV.}
\label{tab:dataset urls and license}
\end{table*}


\end{document}